\documentclass[11pt,a4paper]{article}
\usepackage{times,latexsym}
\usepackage{url}
\usepackage[T1]{fontenc}
\usepackage{multirow}
\usepackage[graphicx]{realboxes}
\usepackage[acceptedWithA]{tacl2018v2}
\usepackage{xspace,mfirstuc,tabulary}

\newif\iftaclinstructions
\taclinstructionsfalse % AUTHORS: do NOT set this to true
\iftaclinstructions
\renewcommand{\confidential}{}
\renewcommand{\anonsubtext}{(No author info supplied here, for consistency with
TACL-submission anonymization requirements)}
\newcommand{\instr}
\fi

\iftaclpubformat % this "if" is set by the choice of options

\else

\fi

\title{What Helps Transformers Recognize Conversational Structure? Importance of Context, Punctuation, and Labels in Dialog Act Recognition}

\author{
 Piotr Żelasko\textsuperscript{\textdagger\textdaggerdbl}, Raghavendra Pappagari\textsuperscript{\textdagger}, Najim Dehak\textsuperscript{\textdagger\textdaggerdbl} \\
 \textsuperscript{\textdagger}Center of Language and Speech Processing, \\
 \textsuperscript{\textdaggerdbl}Human Language Technology Center of Excellence, Johns Hopkins University, Baltimore, MD, USA \\
  {\sf piotr.andrzej.zelasko@gmail.com} \\
}

\date{}

\begin{document}
\maketitle
\begin{abstract}
Dialog acts can be interpreted as the atomic units of a conversation, more fine-grained than utterances, characterized by a specific communicative function. 
The ability to structure a conversational transcript as a sequence of dialog acts -- dialog act recognition, including the segmentation -- is critical for understanding dialog.
We apply two pre-trained transformer models, XLNet and Longformer, to this task in English and achieve strong results on Switchboard Dialog Act and Meeting Recorder Dialog Act corpora with dialog act segmentation error rates (DSER) of 8.4\% and 14.2\%.
To understand the key factors affecting dialog act recognition, we perform a comparative analysis of models trained under different conditions.
We find that the inclusion of a broader conversational context helps disambiguate many dialog act classes, especially those infrequent in the training data.
The presence of punctuation in the transcripts has a massive effect on the models' performance, and a detailed analysis reveals specific segmentation patterns observed in its absence.
Finally, we find that the label set specificity does not affect dialog act segmentation performance.
These findings have significant practical implications for spoken language understanding applications that depend heavily on a good-quality segmentation being available.
\end{abstract}

\section{Introduction}
\label{sec:introduction}

\begin{figure*}[t]
    \centering
    \includegraphics[width=\linewidth]{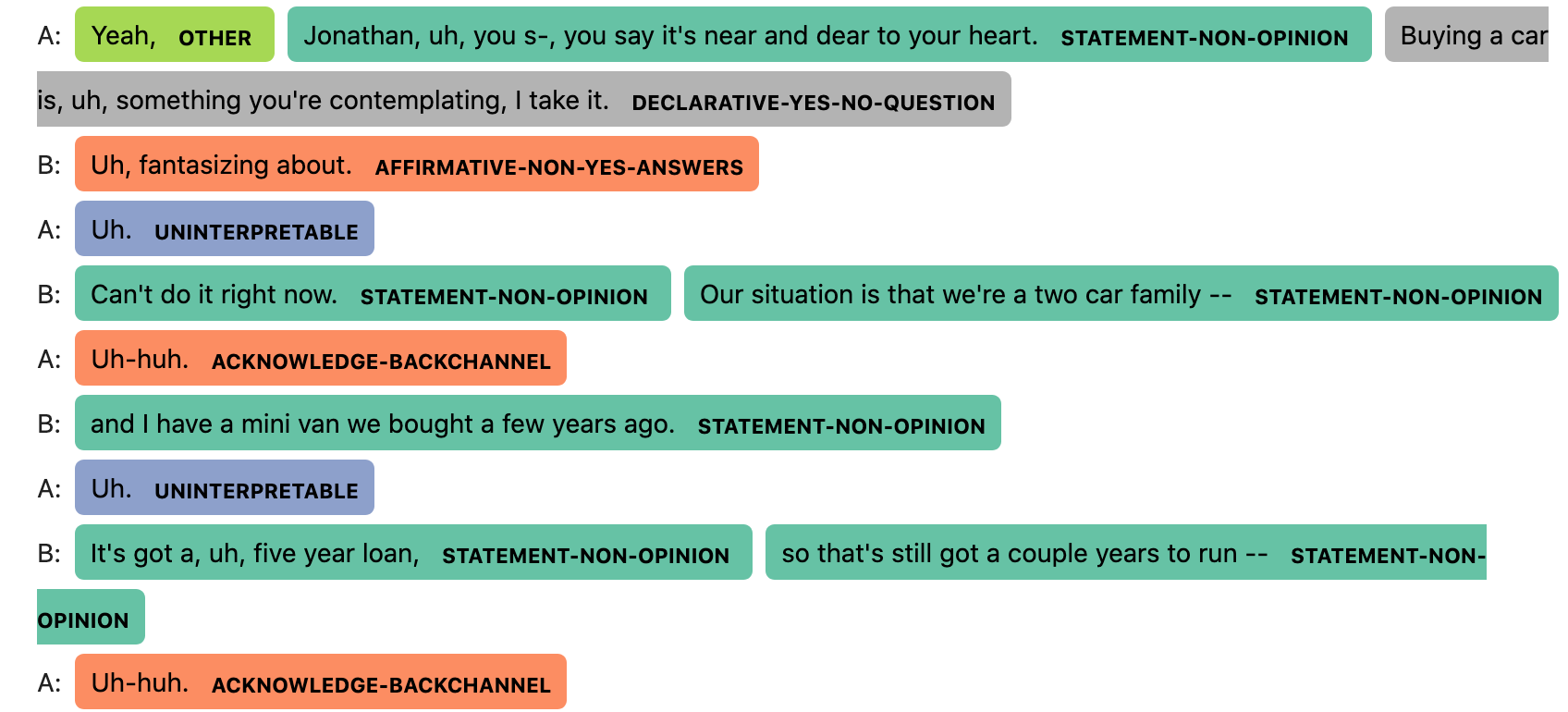}
    \caption{An illustration of dialog acts in a Switchboard conversation. Note how the speaker turns may consist of multiple dialog acts, indicating a different function for each utterance. Dialog act annotation allows us to segment the conversation into meaningful units that can be used for downstream processing in spoken language understanding (SLU) applications.}
    \label{fig:dialog_act_illustration}
\end{figure*}

% What is the main focus of the paper

The human dialog is a never-ending source of diversity, abundant with exceptions and surprising ways to express one's thoughts. As a community, we have spent a massive effort in the past few decades to help the machine achieve even the slightest level of understanding of our means of communication. Remarkably, to some extent, we have succeeded. A consequence of this fact is the widespread presence of so-called \textit{voice assistants}, i.e., conversational agents of limited capabilities, which have gained much popularity in recent years. 

While the main focus of modern dialog research is placed on these human-machine interactions, it is the conversation between humans that poses the greatest challenges to spoken language understanding.
Consider the task of intent recognition -- in a goal-oriented dialog, where the human expects their machine interlocutor to have only limited understanding capabilities, one can reasonably expect there to be a single, self-contained and straightforward utterance expressing the person's request. \citet{siegert2018we} show in a subjective evaluation of Alexa users that they consider such a conversation "more difficult" than talking to a human. With a simpler dialog structure, it is natural to approach intent recognition as a multi-class classification task, by classifying each utterance's underlying intent. 

The same task of intent recognition becomes much more complex when the dialog involves two or more humans. 
Their conversations are riddled with various disfluencies, such as discourse markers, filled pauses, or back-channeling~\cite{charniak2001edit}. \citet{shalyminov2018multi} propose multi-task training for a disfluency detection model capable of spotting hesitations, prepositional phrase restarts, clausal restarts, and corrections.
Spontaneous dialogs are also characterized by much more dynamic structure than written text data.
\citet{kempson2000dynamic,kempson2016language} show that dialog may be viewed as a sequence of incremental contributions -- called \emph{split utterances} -- rather than complete sentences, and propose the \emph{Dynamic Syntax} paradigm, claiming that standard syntactic models are insufficient to capture dialog.
Another study~\cite{purver-etal-2009-split} finds that up to 20\% of utterances in the British National Corpus~\cite{burnard2000british} dialogs fit the definition of \emph{split utterances}, with about 3\% of them being cross-speaker utterance completions.
\citet{eshghi-etal-2015-feedback} propose to view backchannels and other discourse markers as feedback in conversation that is a core component of its semantic structure, rather than a nuisance in downstream processing.
This point is further argued by \citet{purver2018computational}, who propose incremental models for detecting miscommunication phenomena in human-human conversations.
Clearly, an attempt to determine a person's intent grows beyond a turn-level classification task in such scenarios.

% What is the problem?

Dialog acts are vital to understanding the structure of the dialog. 
One of their modern definitions states that they are atomic units of conversation, which are more fine-grained than utterances and more specific in their function~\cite{pareti2018dialog}. The part of utterance that forms a dialog act is also known as a functional segment.
Recently, the definition, taxonomy, and annotation process of dialog acts has been standardized through an ISO norm~\cite{bunt2012iso,bunt2017revisiting,bunt2020iso}.
Earlier studies on this topic typically used custom-tailored dialog act sets -- notably, this category includes the Dialog Act Markup in Several Layers (DAMSL) scheme~\cite{core1997coding}, which was later adopted and modified to annotate the Switchboard corpus~\cite{jurafsky1997switchboard,stolcke2000dialogue}.
Interestingly, dialog acts are related to the philosophy of language \emph{speech acts} theory introduced initially by \citet{austin1962}, in the sense that they view utterances as actions performed by the speakers.

% Why is the problem hard? Why hasn't it been solved before?

Dialog act recognition typically entails two tasks: dialog act segmentation (DAS) and dialog act classification (DAC).
In this work, we address both of them jointly and refer to their combination further as dialog act recognition.
At the time of the conception of the first widely studied corpus for this task, the Switchboard Dialog Act (SWDA), DAS was considered a problem too difficult to address, and the pioneering works focused solely on the classification of dialog acts given the oracle segmentation~\cite{stolcke2000dialogue}. More recent works attempt to retrieve the segmentation through conditional random fields (CRF) or recurrent neural networks (RNN). However, these models still suffer from a significant margin of error, as shown by~\citet{zhao2019joint} and later in Section~\ref{sec:results:turn_level}.
It is worth noting that in some downstream applications, the availability of high-quality segmentation is valuable regardless of any classification errors: some examples include intent classification~\cite{pareti2018dialog}, semantic clustering~\cite{bergstrom2009conversation}, or temporal sentiment analysis~\cite{clavel2015sentiment}, all of which heavily depend on the segmentation.

To the best of our knowledge, the DAS performance of transformer models~\cite{vaswani2017attention} has not yet been investigated. Transformers recently demonstrated state-of-the-art performance across a range of natural language processing (NLP) tasks when combined with language model pre-training~\cite{devlin2018bert,yang2019xlnet,liu2019roberta,beltagy2020longformer}. A major obstacle in applying transformer models to DAS is their $O(n^2)$ computational complexity w.r.t. the input sequence length, making it infeasible to process conversations longer than a couple of hundred tokens. Thus, there are few transformers applications to segmentation tasks -- e.g.,~\citet{Glavas2020TwoLevelTA} employed transformers for topic segmentation, but they assume that text had already been segmented and uses the sentence representations instead of word representations as input to transformers.

% What are key components of our proposal?

To address the transformers' limitations, we investigate two approaches. In the first one, we use XLNet~\cite{yang2019xlnet}, a model based on the TransformerXL architecture~\cite{dai2019transformer}, which is capable of processing the input sequence in windows while propagating the activations of the intermediate layers across as additional inputs in the following window. In the second approach, we use Longformer~\cite{beltagy2020longformer}, which processes the whole sequence in a single pass, but for each token attends only to neighboring $N$ other tokens, reducing the complexity to $O(mn)$, which is linear w.r.t. the input length.

Furthermore, we ask several questions to understand better the factors affecting dialog act recognition and design the experiments accordingly:
\begin{itemize}
    \item \emph{What is the significance of seeing a larger context in dialog act recognition?} Contextual dialog act models have been considered before, but they were either classification models with oracle segmentation or segmentation models that look at a limited number of past turns (see Sections~\ref{sec:related_work:classification} and~\ref{sec:related_work:segmentation}).
    \item \emph{How strongly does text formatting, i.e., the presence of punctuation and capitalization, affect the segmentation quality?} This question is of significant practical importance -- speech transcripts are often obtained through an automatic speech recognition (ASR) system, and many of them do not offer enhanced text formatting capabilities.
    \item \emph{How do the size and the specificity of the dialog act label set affect the recognition difficulty?} In some applications, the segmentation itself might be more important than having a dialog act label -- e.g., when clustering utterances to discover the expressions with similar meaning. Would a large, detailed dialog act label set still be beneficial for such scenarios? Are dialog act labels necessary at all, or is it sufficient to know when they begin and end?
\end{itemize}

\section{Related work}
\label{sec:related_work}

\subsection{Switchboard dialog act}
\label{sec:related_work:swda}

The most widely studied dialog act dataset is Switchboard (SWDA)~\cite{jurafsky1997switchboard,jurafsky1998johns}. It consists of telephone conversations, first manually segmented into turns and \textit{utterances} -- later formally called \emph{functional segments}~\cite{bunt2012iso}, i.e. the units of dialogue act annotation. \citet{bunt2012iso} define them as a minimal stretch of behavior with one or more communicative functions. The total word count is about 1.4M. The conversations have 1454 words on average, and the longest one has 3122 words. The Switchboard annotators originally used the DAMSL labeling scheme~\cite{core1997coding} with 220 dialog acts and clustered them after annotation into a reduced label set. There seems to be no consensus on the reduced label set size -- some of the works using a 42 labels set are~\citet{quarteroni2011simultaneous,liu-etal-2017-using-context,ortega2018lexico,kumar2018dialogue} -- others use a 43 labels set~\cite{ortega2017neural,raheja2019dialogue,zhao2019joint,dang2020end}.

\subsection{Meeting recorder dialog act}
\label{sec:related_work:mrda}

Meeting Recorder Dialog Act (MRDA)~\cite{shriberg2004icsi} is a corpus of 75 meetings that took place in the International Computer Science Institute (ICSI). The conversations involve more than two speakers and are significantly longer than those in SWDA. The mean word count is about 11k, and the longest dialog has 22.5k words. There are 850k words in total, making MRDA approximately half the size of SWDA. The dialog act labeling scheme is different from that in SWDA - the annotators used a 51 act set that significantly overlaps with SWDA-DAMSL (we refer to that as the \emph{full} set). These acts were later clustered, with two granularity levels, into a \emph{general} set of 12 acts and a \emph{basic} set of 5 acts. The \emph{basic} set is reduced to the following classes: \emph{Statement}, \emph{Question}, \emph{Backchannel}, \emph{Disruption}, and \emph{Floor-Grabber}. We refer the reader to~\citet{shriberg2004icsi} for a detailed comparison of dialog act classes between SWDA and MRDA.

\subsection{Dialog act classification}
\label{sec:related_work:classification}

There are two main groups of studies: the first assumes that the segmentation is known and considers dialog act recognition as a pure classification task. The original SWDA authors first take such an approach with a hidden Markov model (HMM)~\cite{jurafsky1998johns}. Others have introduced conditional random fields (CRF) to solve this task~\cite{quarteroni2011simultaneous}. Some authors found that considering the context explicitly in RNN models helps dialog act classification~\cite{ortega2017neural,liu-etal-2017-using-context,kumar2018dialogue,raheja2019dialogue,dai2020local}. Also, it has been shown that incorporating acoustic/prosodic features helps as well to some extent~\cite{ortega2018lexico,si2020hierarchical}. \citet{colombo2020guiding} report the best result to date for SWDA classification - an accuracy of 85\%, obtained by a sequence-to-sequence (seq2seq) GRU model with guided attention. For MRDA, the best classification accuracy is 92.2\% reported by~\cite{li2019dual}, achieved with a dual-attention hierarchical BiGRU with a CRF on top. These approaches are not directly comparable with ours, as they assume an oracle segmentation of the transcript.

\subsection{Dialog act segmentation and recognition}
\label{sec:related_work:segmentation}

More interesting in the context of our work are the studies that consider dialog act segmentation and recognition. One of the first attempts has been made by \citet{ang2005automatic} with decision trees and HMMs for the MRDA corpus. CRF has been successfully employed in this task~\cite{quarteroni2011simultaneous}. 
The closest work to ours is by \citet{zhao2019joint}, where a bidirectional gated recurrent unit (BiGRU) model is used to segment and classify dialog acts in SWDA jointly. The model is considered as a sequence tagger with an optional CRF layer or in an encoder-decoder setup. It also integrates previous dialog act predictions for ten previous turns using an attention mechanism. Notably, the main differences from our setup are that \citet{zhao2019joint}:
\begin{enumerate}
    \item consider prediction for a single turn at a time, whereas our dialog-level contextual models process multiple turns at the same time, which allows to include both past and future context into prediction;
    \item use exclusively lowercase text without punctuation, whereas we study setups both with and without the punctuation and truecasing;
    \item limit the vocabulary at 10000 words, whereas we use sub-word tokenizers with no such limitation -- this results in the model being able to leverage another 10000 less frequent words in SWDA, that would have otherwise been replaced by an out-of-vocabulary symbol;
    \item connect dialog act continuations (the segments labeled in SWDA with a $+$) to the previous turn when interrupted, e.g., by a backchannel -- we view that operation as a work-around for their models to be able to see the relevant future context, whereas our proposed models require no such pre-processing.
\end{enumerate}
Finally, we provide a more detailed analysis of the effect of context on the recognition outputs; we also investigate the effect of punctuation and label set specificity, which is not discussed in that work.
   
\subsection{The effect of context and punctuation}
\label{sec:related_work:context}

In~\cite{liu2017using}, the authors process each dialog act segment in parallel streams using a CNN and combine the sequence of sentence representations using an LSTM to exploit the context.
The influence of context is explored in~\cite{bothe2018context} by using an LSTM on the segment representations.
Here, dialog act classification is achieved in two stages: learning segment representations and dialog act classification using an LSTM.
The usage of punctuation marks as features and other heuristics, such as the number of words in the segment, n-grams, the dialog act of the next segment, and others, is explored in~\cite{samuel1998dialogue, verbree2006dialogue}. However, the effect of each of these heuristics, especially punctuation marks, is not analyzed.
To the best of our knowledge, there are no studies that attempt to understand the role of context, punctuation, or label set specificity on dialog act recognition in-depth.

\section{Methods}
\label{sec:methods}

\subsection{Transformers}
The transformer architecture is shown to produce state-of-the-art results on several NLP tasks~\cite{vaswani2017attention,devlin2018bert}.
It consists of repeated blocks of a self-attention layer and a feed-forward layer.
The self-attention layer processes the entire input sequence and learns to attend to the relevant tokens by computing the cross-token similarity in the input sequence.
The similarity computation is implemented with a dot-product followed by a softmax operation.
Each token's representation in the self-attention layer output is passed through a feed-forward layer before the next self-attention layer.
However, as the self-attention layer processes all tokens of the input sequence simultaneously, it is invariant to the input sequence's token order.
The ordering information is preserved by adding positional embeddings to the input token embeddings.
Positional embeddings include one vector per token position and are learned during model training together with other model parameters.

One major limitation of transformer models is their scalability to longer inputs, as the complexity of each self-attention layer is $O(n^2)$ where $n$ is the input sequence length.
More recent works address this limitation in several ways: 1) propagation of context between segments of long sequence~\cite{dai2019transformer, yang2019xlnet}, 2) local attention~\cite{ye2019bp,beltagy2020longformer, wu2020lite, zaheer2020bigbird}, 3) sparse attention~\cite{kitaev2020reformer, tay2020sparse, zaheer2020bigbird}, 4) efficient attention operation~\cite{wang2020linformer, katharopoulos2020transformers,shen2021efficient}.
In this work, we explore two of these models for dialog act recognition: XLNet~\cite{yang2019xlnet} which is based on the propagation of context, and Longformer~\cite{beltagy2020longformer} which uses local attention.

\subsection{XLNet}
XLNet~\cite{yang2019xlnet} is a transformer model trained with a masked language model (MLM) criterion. It consists of 12 (\emph{base}) or 24 (\emph{large}) self-attention layers. It is based on TransformerXL~\cite{dai2019transformer}, which enables it to process text sequences in windows while propagating the context in the forward direction. We leverage this property to process conversational transcripts efficiently. Furthermore, XLNet is pre-trained as an autoregressive language model that maximizes the expected likelihood over all permutations of the input sequence factorization order.
It is interesting to note that this model, unlike BERT, uses relative positional encodings that do not need to be learned, making it possible to process sequences of arbitrary lengths. Even then, the quadratic computational complexity necessarily renders such processing infeasible, making windowed processing a more practical choice.

\subsection{Longformer}
Longformer~\cite{beltagy2020longformer} is based on a modification of the self-attention layer that reduces the computational complexity by limiting the context available to each input token. It splits the attention into two components - local and global. The local component is a sliding window of fixed size for each self-attention layer, dramatically reducing long sequences' computational complexity. The global component allows select tokens to attend to the entire sequence. We do not use it in this work -- unlike in text classification, where [CLS] uses global attention, or question answering, where the question tokens use global attention~\cite{beltagy2020longformer}, there are no clear candidates for it in dialog act recognition.
Following \citet{beltagy2020longformer}, we use RoBERTa~\cite{liu2019roberta} -- BERT with carefully tuned hyperparameters -- as the base model to avoid the costly pre-training process. This model's limitation is that it cannot process token sequences longer than those seen during training (4096 tokens for the pre-trained model open-sourced by~\citet{beltagy2020longformer}). We investigate Longformer because we consider its sliding window attention mechanism as a natural extension over the XLNet's window-processing mechanism.

\section{Experimental setup}
\label{sec:experimental_setup}

\subsection{Model training}
\label{sec:experimental_setup:model_training}

For both transformer models, we use pre-trained sub-word tokenizers and weights, as provided by HuggingFace\footnote{\url{https://huggingface.co/}} -- \emph{allenai/longformer-base-4096} for Longformer and \emph{xlnet-base-cased} for XLNet. These are the \textit{base} variants with 12 self-attention layers. To adapt the models to the DAS task, we put a token classification layer on top of the transformer and train it with a per-token cross-entropy loss. We fine-tune each model on the training portion of the dataset - 1003 calls for SWDA and 51 meetings for MRDA. We use the validation set (112 SWDA calls; 12 MRDA meetings) to select the best model for each variant and the test set (19 SWDA calls; 12 MRDA meetings) for the final evaluation.

The baseline BiGRU model is trained in the same setup as described in~\citet{zhao2019joint}. For both XLNet and Longformer, we compare their performance to BiGRU by training them as turn-level models that see only a single speaker turn without additional context. In a separate experiment, to measure the effect of providing the surrounding dialog context, we train them as broad-context models processing either full transcripts (Longformer) or chunks (XLNet). All reported metrics are the mean values from three runs with different random seeds (42, 43, 44).

We train each model with a single GeForce GTX 1080 Ti GPU, which allowed us to construct batches of 6 chunks with 512 tokens each for XLNet training. The same setup might not be optimal for Longformer, as only the first 512 positional embeddings would have been fine-tuned. Therefore, we train it with 4096 token windows and an effective batch size of 6, using gradient accumulation. All models are trained for ten epochs with an Adam optimizer, a learning rate of 5e-5, and a learning schedule linearly decreasing its value towards 0. We evaluate the model on the validation set after each epoch and select the model that achieved the best F1 macro score to report the test set results.

\subsection{Data preparation}
\label{sec:experimental_setup:data_preparation}

To transform the SWDA\footnote{We use the SWDA distribution available here: \url{http://compprag.christopherpotts.net/swda.html}} and MRDA\footnote{We use the MRDA distribution available here: \url{https://github.com/NathanDuran/MRDA-Corpus}} conversational transcripts into model inputs, we perform several steps. First, we remove all annotator comments from the SWDA text. We evaluate each model in two variants - with/without punctuation and truecasing, to investigate how strongly it affects the performance. When punctuation and truecasing are used, they are always the ground truth. To create a single sequence out of speaker turns, we concatenate them with a unique \emph{TURN} token in between that does not participate in loss computation but explicitly indicates that the speaker has changed.

Following~\citet{zhao2019joint}, we encode the dialog act labels using an \textbf{E} joint coding scheme. In the \textbf{E} scheme, each word comprising a dialog act is assigned a label - the \textbf{E} label indicates an end of the dialog act, and the \textbf{I} label indicates a token other than an ending. The joint coding also specializes the \textbf{E} label for each dialog act class in the label set, allowing to perform dialog act recognition. The \textbf{I} label is shared between all dialog act classes. BERT models typically use sub-word tokenization - byte-pair encoding~\cite{gage1994new,sennrich2015neural} for Longformer and SentencePiece~\cite{kudo2018sentencepiece} for XLNet. When a word is split into multiple tokens, we assign the dialog act label only to the first token and discard the following tokens' predictions (i.e., they do not participate in loss computation and are ignored when reading predictions during inference).

For SWDA, we are using the 42 dialog act labels (as \emph{Abandoned-or-Turn-Exit} act is merged with \emph{Uninterpretable}) encoded into 43 labels in total, including the \textbf{I} label. We experiment with all the label sets available in MRDA - \emph{basic} with 5 labels, \emph{general} with 12 labels, and \emph{full} with 51 labels (6, 13 and 52 respectively when counting the \textbf{I} label). Unless otherwise specified, we always use the 5 labels set for MRDA and 42 labels for SWDA.

Some SWDA dialog acts are extended across turns with a $+$ label, e.g., when somebody interrupted with a backchannel. We respect that by assigning an \textbf{I} label to the last token in the interrupted turn, thus creating a multi-turn functional segment.

For inference, the calls are processed in sliding windows. With XLNet, we use a window size of 512 tokens without overlap. We compare the predictions with and without the context propagation across windows to understand its importance. With Longformer, we do not need to explicitly construct the windows, as each token's attention is limited to a local context of 256 neighboring tokens on each side.

\subsection{Metrics}
\label{sec:experimental_setup:metrics}

To measure the model performance, we use standard \emph{micro} and \emph{macro} weighted F1 metrics, as well as metrics explicitly evaluating the segmentation quality~\cite{granell2010dialogue,zhao2019joint}:
\begin{itemize}
    \item \emph{Dialog Act Segmentation Error Rate} (DSER) measures the percentage of reference segments that were not recognized with perfect boundaries, disregarding the dialog act label.
    \item \emph{Segmentation Word Error Rate} (SegWER) is additionally weighted by the number of words in a given segment.
    \item \emph{Dialog Act Error Rate} (DER) is computed similarly to DSER but also considers whether the dialog act label is correct.
    \item \emph{Joint Word Error Rate} (JointWER) is a word count weighted version of DER.
\end{itemize}

Note that these metrics are strict: if a 3-word turn with a single \emph{Statement} act is recognized as an \emph{Acknowledgment} on the first word and \emph{Statement} on the next two, the micro F1 score is 66.6\%, the macro F1 score is 55.5\%, but the error rate metrics are all at 100\%.

For reference, when reading the dialog act metrics, the SWDA and MRDA test sets have respectively 4500 and 16702 functional segments. For reading micro and macro F1 scores, SWDA and MRDA test sets have 29.8K and 100.6K words.

% The big tables come here!

\begin{table*}[h!]
    \centering
    \resizebox{\textwidth}{!}{
    \begin{tabular}{lllrrrrrr}
    \hline
    \textbf{Case} & \textbf{Dataset} & \textbf{Model} &  \textbf{micro\_f1} &  \textbf{macro\_f1} &   \textbf{DSER} &  \textbf{SegWER} &    \textbf{DER} &  \textbf{JointWER} \\
    \hline\hline
    lower & MRDA & BiGRU        &     92.66 &     64.68 &  41.69 &   51.56 &  54.78 &     59.54 \\
            &      & Longformer &     \textbf{94.02} &     \textbf{70.25} &  34.55 &   41.15 &  45.74 &     46.71 \\
            &      & XLNet      &     \textbf{94.02} &     69.54 &  \textbf{33.62} &   \textbf{40.40} &  \textbf{45.62} &     \textbf{46.38} \\
            \cline{2-9}
            & SWDA & BiGRU      &     92.90 &     34.16 &  29.31 &   40.51 &  49.59 &     57.83 \\
            &      & Longformer &     \textbf{94.04} &     \textbf{41.15} &  20.27 &   28.50 &  \textbf{40.29} &     45.45 \\
            &      & XLNet      &     93.99 &     39.56 &  \textbf{19.79} &   \textbf{27.12} &  41.13 &     \textbf{45.18} \\
    \hline\hline
    nolower & MRDA & BiGRU      &     96.60 &     79.21 &  18.28 &   22.31 &  27.91 &     25.67 \\
            &      & Longformer &     97.08 &     80.80 &  16.19 &   18.05 &  25.34 &     20.26 \\
            &      & XLnet      &     \textbf{97.12} &     \textbf{81.71} &  \textbf{15.08} &   \textbf{17.81} &  \textbf{24.01} &     \textbf{19.89} \\
            \cline{2-9}
            & SWDA & BiGRU      &     94.47 &     38.92 &  14.21 &   22.31 &  37.86 &     44.62 \\
            &      & Longformer &     95.35 &     \textbf{46.87} &  11.00 &   16.21 &  32.31 &     35.78 \\
            &      & XLnet      &     \textbf{95.40} &     46.24 &   \textbf{9.98} &   \textbf{14.64} &  \textbf{31.85} &     \textbf{34.67} \\
    \hline
    \end{tabular}
    }
    \caption{\label{tab:results_turn_level}
    Dialog act recognition performance for BiGRU (baseline), XLNet, and Longformer models on SWDA and MRDA datasets. The models are processing each speaker turn separately, without seeing any additional context.
    }
\end{table*}

\begin{table*}[h!]
    \centering
    \resizebox{\textwidth}{!}{
    \begin{tabular}{lllrrrrrr}
    \hline
    \textbf{Case} & \textbf{Dataset} & \textbf{Model} & \textbf{micro\_f1} &  \textbf{macro\_f1} &   \textbf{DSER} &  \textbf{SegWER} &    \textbf{DER} &  \textbf{JointWER} \\
    \hline\hline
    lower & MRDA   & Turns$\dagger$      &     94.02 &     69.54 &  33.62 &   40.40 &  45.62 &     46.38 \\
            &      & Longformer          &     94.65 &     75.30 &  32.78 &   39.70 &  44.11 &     45.17 \\
            &      & XLNet               &     94.82 &     75.49 &  \textbf{32.71} &   38.74 &  43.78 &     44.21 \\
            &      & \quad\textit{+prop} &     \textbf{94.89} &     \textbf{75.82} &  32.87 &   \textbf{38.32} &  \textbf{43.61} &     \textbf{43.76} \\
            \cline{2-9}
            & SWDA & Turns$\dagger$      &     93.99 &     39.56 &  19.79 &   27.12 &  41.13 &     45.18 \\
            &      & Longformer          &     95.51 &     53.70 &  18.60 &   25.17 &  38.60 &     45.55 \\
            &      & XLNet               &     95.49 &     53.48 &  17.74 &   24.24 &  37.99 &     44.88 \\
            &      & \quad\textit{+prop} &     \textbf{95.57} &     \textbf{54.86} &  \textbf{17.48} &   \textbf{24.09} &  \textbf{37.51} &     \textbf{44.38} \\
    \hline\hline
    nolower & MRDA & Turns$\dagger$      &     97.12 &     81.71 &  15.08 &   17.81 &  24.01 &     19.89 \\
            &      & Longformer          &     97.45 &     85.31 &  14.52 &   17.41 &  22.87 &     19.45 \\
            &      & XLNet               &     \textbf{97.57} &     85.54 &  14.43 &   \textbf{16.59} &  22.56 &     \textbf{18.59} \\
            &      & \quad\textit{+prop} &     97.55 &     \textbf{85.67} &  \textbf{14.15} &   16.85 &  \textbf{22.29} &     18.92 \\
            \cline{2-9}
            & SWDA & Turns$\dagger$      &     95.40 &     46.24 &   9.98 &   14.64 &  31.85 &     \textbf{34.67} \\
            &      & Longformer          &     96.58 &     57.73 &   8.76 &   12.98 &  30.73 &     36.41 \\
            &      & XLNet               &     96.57 &     57.91 &   8.40 &   \textbf{12.28} &  30.67 &     36.42 \\
            &      & \quad\textit{+prop} &     \textbf{96.65} &     \textbf{58.17} &   \textbf{8.39} &   12.34 &  \textbf{30.21} &     35.90 \\
    \hline
    \end{tabular}
    }
    \caption{\label{tab:results_dialog_level}
    Dialog act recognition performance of large-context models -- Longformer and XLNet. XLNet\emph{+prop} means that the intermediate activations are passed between the processed segments during inference. $\dagger$ -- The best turn-level model, i.e., the XLNet, is used as a baseline -- \emph{Turns}.
    }
\end{table*}

\begin{table*}
    \centering
    \begin{tabular}{lllrrrrrr}
\hline
\textbf{Case} & \textbf{Dataset} & \textbf{Tagset} & \textbf{micro\_f1} &  \textbf{macro\_f1} &   \textbf{DSER} &  \textbf{SegWER} &    \textbf{DER} &  \textbf{JointWER} \\
\hline\hline
lower   & MRDA & 51     &     91.90 &     30.94 &  32.93 &            39.15 &  58.62 &     63.90 \\
        &      & 12     &     94.07 &     48.39 &  35.51 &            40.56 &  48.72 &     49.42 \\
        &      & 5$^*$  &     94.89 &     75.82 &  32.87 &            \textbf{38.32} &  43.61 &     43.76 \\
        &      & 1      &     96.74 &     95.23 &  \textbf{32.85} &            38.94 &  -     &     -     \\
\cline{2-9}       
        & SWDA & 42$^*$ &     95.57 &     54.86 &  \textbf{17.48} &            \textbf{24.09} &  37.51 &     44.38 \\
        &      & 1      &     98.20 &     97.45 &  17.51 &            24.32 &  -     &     -     \\
\hline\hline
nolower & MRDA & 51     &     93.85 &     40.65 &  \textbf{13.88} &            17.38 &  45.22 &     49.11 \\
        &      & 12     &     96.57 &     64.51 &  14.21 &            17.42 &  27.62 &     26.96 \\
        &      & 5$^*$  &     97.55 &     85.67 &  14.15 &            16.85 &  22.29 &     18.92 \\
        &      & 1      &     98.76 &     98.21 &  14.55 &            \textbf{16.52} &  -     &     -     \\
\cline{2-9}
        & SWDA & 42$^*$ &     96.65 &     58.17 &   8.39 &            12.34 &  30.21 &     35.90 \\
        &      & 1      &     99.22 &     98.89 &   \textbf{8.37} &            \textbf{12.18} &  -     &     -     \\
\hline
\end{tabular}
    \caption{\label{tab:label_set_comparison}
    \emph{XLNet+prop} segmentation and recognition results for different label sets granularities; in MRDA: \emph{full} (51), \emph{general} (11), \emph{basic} (5), and pure segmentation (1); in SWDA \emph{basic} (42) and pure segmentation (1). DER and JointWER are not defined for pure segmentation. All experiments are performed using full dialog context, with identical hyperparameters, except for the output layer size. The asterisk denotes the label sets typically used in other works.
    }
\end{table*}

\section{Results}
\label{sec:results}

In this section, we present the results of our experimental evaluation. Each result table is first split into \textit{lower} and \textit{nolower} sections, which respectively stand for a lowercase transcript with no punctuation, and an original case transcript with punctuation symbols. For both scenarios, we always show the results on both MRDA and SWDA datasets.

\subsection{Single turn context models}
\label{sec:results:turn_level}

We start our experiments by investigating how much improvement we can achieve by replacing a simple but established BiGRU baseline model with one of the transformer models. The baseline is trained in the same setup as in~\citet{zhao2019joint}\footnote{During replication, we discovered an issue in the experimental results reported in that paper - the segment insertion errors were not counted, which artificially lowered the error rates. We contacted the authors and agreed that the results we report for their model are the correct ones.}. To make the comparison fair, we train the XLNet and Longformer on single turn inputs so that the model does not see any dialog context. The same is true during inference. The results are shown in Table~\ref{tab:results_turn_level}. 

Both transformer models offer substantial improvements over the BiGRU baseline in all scenarios. In most evaluations, XLNet achieves the best results, outperforming Longformer by a small margin, compared to the improvement over BiGRU. Since these experiments do not test the model's ability to handle long-range context, these results suggest that XLNet's pre-training procedure is more suitable for dialog act recognition than that of Longformer.

\subsection{Broad context models}
\label{sec:results:dialog_level}

In the second experiment, we investigate how long-document transformers perform in dialog act recognition. As a baseline (\textit{Turns}), we re-use the best model from Section~\ref{sec:results:turn_level} -- XLNet -- processing dialog transcript on a turn-by-turn basis without additional context. The other proposed models process the whole transcript in sliding windows. XLNet uses a window of 512 tokens with a step size of 512 tokens. This window traversal strategy is not optimal - the tokens on the window boundaries cannot attend to other tokens close by but belonging to another window. XLNet\emph{+prop} partially addresses this issue by propagating the intermediate activations between the windows. Longformer uses a window of 512 tokens with a step size of 1 token, which is possible thanks to its special local attention pattern. Therefore, it fully avoids XLNet's traversal strategy issue. The results are in Table~\ref{tab:results_dialog_level}.

All broad context models outperform the turn-level baseline across all metrics, except the turn-level SWDA \textit{nolower} baseline in the JointWER metric. XLNet\textit{+prop} emerges as the best model in all configurations with minor gains over XLNet. Similarly, as in Section~\ref{sec:results:turn_level}, we observe consistent improvements in all setups when using XLNet instead of Longformer. However, we cannot conclude that XLNet uses the context more effectively, as its performance on context-less turn prediction was also better than that of  Longformer's. Besides the attention patterns, there are other differences between the models, such as the pretraining conditions and positional encoding schemes, which could also explain the observed results. However, it is an indication that limiting Longformer's number of positional embeddings to 4096 is not a limiting factor in its performance.

\begin{figure*}[h!]
    \centering
    \resizebox{\textwidth}{!}{\includegraphics{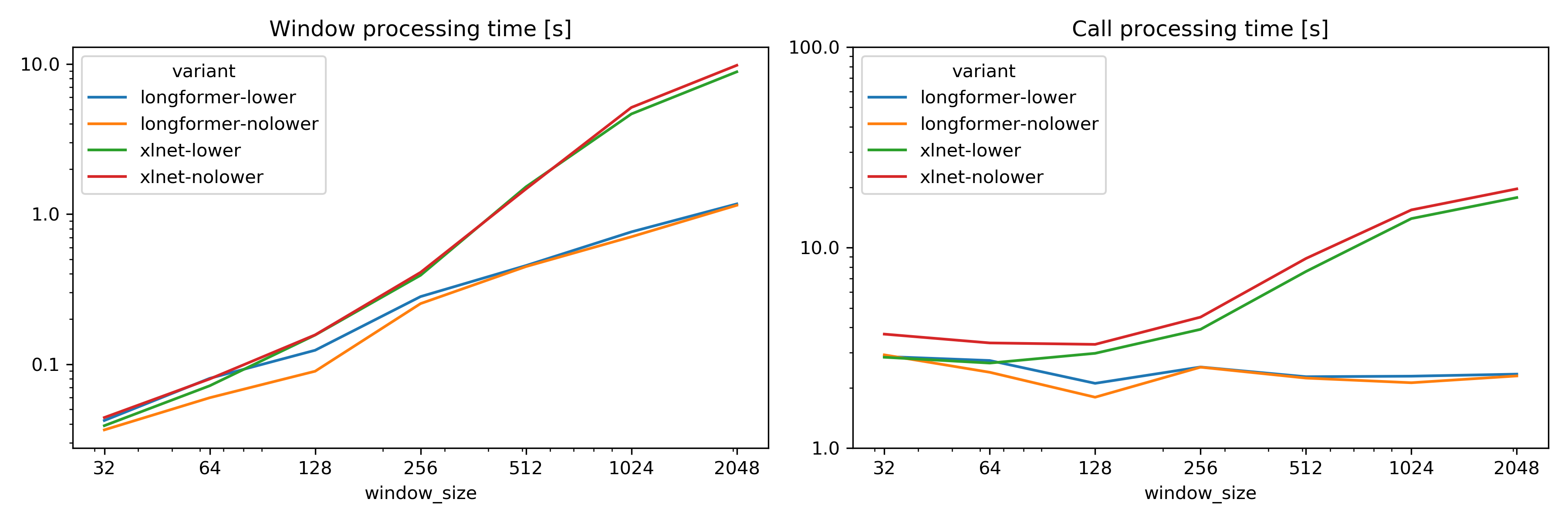}}
    \caption{Prediction time for SWDA call \emph{sw2229} by Longformer and XLNet with different window sizes. The left-side plot shows the mean time it takes to predict a single window, and the right-side plot shows the time needed to process the full dialog. Window sizes larger than 512 imply sub-windowing for Longformer, which in this experiment has learned only 512 positional embeddings.}
    \label{fig:results:processing_times}
\end{figure*}

We compare the runtime of XLNet and Longformer models. Average inference time with 512 tokens window on SWDA transcripts with an eight-core Intel Core i9-9980HK CPU takes 2.8 seconds for Longformer and 14.7 seconds for XLNet, making Longformer about five times faster when deployed on a CPU. Figure~\ref{fig:results:processing_times} shows the time it takes for dialog act prediction on a 1750 words call \emph{sw2229} from SWDA -- for smaller windows of 32 and 64, the models take similar time to run, but as the window size increases, Longformer becomes quicker than XLNet. To summarize, Longformer might be more suitable for practical applications, even if it achieves slightly worse recognition results.

An analysis of confusion patterns in the most performant model (\textit{nolower} XLNet\textit{+prop}) does not reveal any new insights in SWDA compared to past works - the most confused label pair is \textit{Statement-opinion} and \textit{Statement-non-opinion}. For the same model in MRDA, we observe the \textit{Question} label has the highest F-score of 98.32\%, followed by 94.38\% for \textit{Statements}. \textit{Backchannels} are the most confused label, with 17\% of them being classified as \textit{Statements}, and 19\% of predicted \textit{Backchannels} being in fact \textit{Statements}. Also, a significant portion of \textit{Disruptions} (25\%) and \textit{Floor-grabbers} (28\%) are confused with the \textit{I} label and respectively 20\% and 14\% of them are predicted as an \textit{I} label. This indicates that these dialog acts are the most difficult to segment correctly - which might be due to only 66.5\% average inter-annotator agreement on MRDA segmentation~\cite{shriberg2004icsi}. Lastly, 13\% of predicted \textit{Floor-grabbers} are in fact \textit{Disruptions}. 

\section{Discussion}

This section presents a detailed analysis of various factors affecting dialog act segmentation and recognition performance. In particular, we look into the effects of label set specificity, punctuation, and context. 

\subsection{The effect of label set specificity}

Since MRDA provides different label set sizes, it is tempting to see how that affects the recognition performance. Furthermore, we investigate a special case where we perform pure segmentation -- i.e., the dialog act labels are stripped, and there remains a single generic \textbf{E} token at the end of each segment. For SWDA, we compare the 42 label set performance with pure segmentation. All experiments are performed using the XLNet\textit{+prop} model, which was the best model in Section~\ref{sec:results:dialog_level}. The results are shown in Table~\ref{tab:label_set_comparison}.

We do not observe a strong effect of the label set size on segmentation performance; the pure segmentation model is practically on par with the dialog act recognition model. This is indicated by little change in DSER and SegWER metrics across the label sets in each experimental scenario. On the other hand, the label set size has a major effect on the classification performance, reflected in F1, DER, and JointWER. We offer two explanations for that. Firstly, the larger label sets have more imbalanced classes, e.g., in the 51 labels set, 43\% of acts are statements, and the 18th most frequent class is already below 1\% of all acts. Secondly, we suspect that the inter-annotator agreement is worse for the large label set, but the MRDA authors only reported it for the five label set (80\% agreement).

\subsection{The effect of dialog context}

\begin{table*}
    \centering
    \resizebox{\textwidth}{!}{
    \begin{tabular}{lrrrr}
    \hline
    \textbf{Mis-segmented dialog acts} &  \textbf{Count} &  \textbf{DSER (turn) [\%]} &  \textbf{DSER (dialog) [\%]} &  \textbf{Abs. gain [\%]} \\
    \hline
    Rhetorical-Questions     &     12 &             58.3 &               16.7 &          -41.7 \\
    Other                    &     15 &             53.3 &               20.0 &          -33.3 \\
    Action-directive         &     30 &             50.0 &               23.3 &          -26.7 \\
    Repeat-phrase            &     21 &             19.0 &                4.8 &          -14.3 \\
    Hedge                    &     23 &             17.4 &                4.3 &          -13.0 \\
    Response-Acknowledgement &     28 &             14.3 &                3.6 &          -10.7 \\
    Statement-non-opinion    &   1494 &             23.0 &               13.7 &           -9.3 \\
    No-answers               &     26 &             19.2 &               11.5 &           -7.7 \\
    Wh-Question              &     56 &             12.5 &                5.4 &           -7.1 \\
    Open-Question            &     16 &              6.2 &                0.0 &           -6.2 \\
    \hline
    \hline
    \textbf{Mis-classified dialog acts} &  \textbf{Count} &  \textbf{DER (turn) [\%]} &  \textbf{DER (dialog) [\%]} &  \textbf{Abs. gain [\%]} \\
    \hline
    Yes-answers                  &     73 &           100.0 &              17.8 &          -82.2 \\
    Open-Question                &     16 &           100.0 &              25.0 &          -75.0 \\
    Repeat-phrase                &     21 &           100.0 &              33.3 &          -66.7 \\
    Wh-Question                  &     56 &            91.1 &              30.4 &          -60.7 \\
    Conventional-closing         &     84 &            65.5 &              10.7 &          -54.8 \\
    Response-Acknowledgement     &     28 &            89.3 &              35.7 &          -53.6 \\
    Rhetorical-Questions         &     12 &           108.3 &              58.3 &          -50.0 \\
    Collaborative-Completion     &     20 &           100.0 &              55.0 &          -45.0 \\
    Backchannel-in-question-form &     21 &            57.1 &              19.0 &          -38.1 \\
    Summarize/reformulate        &     25 &           100.0 &              72.0 &          -28.0 \\
    \hline
    \end{tabular}
    }

    \caption{Top 10 SWDA dialog acts that benefit from dialog-level context availability in pure segmentation and dialog act recognition. The columns denoted by (turn) and (dialog) represent numbers for turn-level XLNet and dialog-level XLNet\emph{+prop}.
    }
    \label{tab:dialog_act_diff}
\end{table*}

To understand how the dialog context helps improve the models, we analyze the predictions of turn-level XLNet and dialog-level XLNet\emph{+prop}. In particular, we find the subset of turns in which the turn-level model made either segmentation or classification errors, but the dialog-level model recognized everything correctly (427 turns, which is 16.3\% of turns in the SWDA test set). This subset contains 752 dialog acts and suffers mostly from misclassification errors — 19.8\% of these dialog acts are mis-segmented with an equal share of over- or under- segmentation, but as many as 75.8\% of them have been misclassified. 

We take a closer look at the differences between the two models' errors by considering the whole test set again and investigating which dialog acts benefitted the most from dialog-level context. To find them, we first have to perform segment-level alignment (since segment boundaries could be misrecognized) using the Levenshtein algorithm. For this purpose, we assume that the reference and predicted segments are equal when they start and end at the same words for pure segmentation and additionally check that their dialog act label is the same for recognition. 

Surprisingly, we find that the strongest turn-level model (XLNet) never correctly recognized more than half of the label set (24 dialog act classes, many of which are infrequent), whereas this number significantly drops for the dialog-level model (4 classes: \emph{Declarative-Wh-Question}, \emph{Dispreferred-answers}, \emph{Self-talk}, \emph{Hold-before-answer-agreement}). The top 10 dialog acts with improved recognition performance, that occurred at least 10 times in SWDA test set, are shown in Table~\ref{tab:dialog_act_diff}. The turn-level model lacked the necessary context to correctly classify \emph{Yes-answers}, \emph{Agree-Accept}, and \emph{Response-Ackonwledgment}, mistaking them mostly for \emph{Ackonwledge-Backchannel}. The model frequently hypothesized \emph{Yes-No-Question} in place of \emph{Wh-Question}. Other highly contextual dialog acts such as \emph{Repeat-phrase}, \emph{Rhetorical-Questions}, \emph{Backchannel-in-question-form} or \emph{Summarize-reformulate} also largely improved.

In terms of segmentation performance differences, the improvements with dialog context are consistent across various kinds of dialog acts: both short (\emph{Response-Acknowledgment}, \emph{No-answers}) and long (\emph{Statement-non-opinion}, \emph{Action-directive}); questions (\emph{Rhetorical-Questions}, \emph{Wh-Question}, \emph{Open-Question}) and statements.

\subsection{The effect of punctuation -- MRDA}

We have previously observed from Table~\ref{tab:label_set_comparison} that removing the capitalization and punctuation has a significant effect on the dialog act recognition.
It suggests a strong correlation between punctuation and dialog acts.
For example, a \textit{Question} dialog act segment might often end with a question mark that could serve as a cue for the model.
In this subsection, we show the correlations between dialog acts and punctuation for MRDA and SWDA datasets.
Table~\ref{tab:mrda_punc_vs_da} presents dialog act vs. punctuation statistics for the MRDA dataset with 5 labels.
Each cell contains the frequency of a dialog act and punctuation occurring together and the percentage of our model errors in parenthesis.

We can observe that the frequency of various punctuation symbols is skewed for each dialog act.
For example, segments with \textit{Statement} and \textit{Backchannel} dialog act labels most often contain full stop, those with \textit{Question} dialog act label contain question mark. 
Similarly, \textit{Floor-grabber} and \textit{Disruption} labelled sentences contain no punctuation.
Given that correlations between dialog acts and punctuation exist, we expect the models to leverage punctuation as a cue for prediction.
Fewer errors (in bold) when punctuation is highly correlated with dialog acts confirm our hypothesis.
For example, dialog act \textit{Question} has a minimal percentage of errors when a question mark is present in the input segment.
Upon further investigation, we found that the ending boundary is consistently recognized correctly when a question mark exists, and any errors that occur are at the segment's beginning.
Also, the high error percentages for dialog acts \textit{Disruption} and \textit{Floor-grabber} could be explained due to their similar distributions of ending punctuation.

\begin{table*}[t!]
    \centering
    %\resizebox{\textwidth}{!}{
    \begin{tabular}{lllll}
    \hline
    {} &    \textbf{Full stop} & \textbf{Excl. mark} & \textbf{Q. mark} & \textbf{None} \\
    \hline
    Backchannel    & \makebox[0.05\textwidth][s]{2120} (\textbf{18.4}) & \makebox[0.02\textwidth][s]{4} (50.0)  & \makebox[0.05\textwidth][s]{0} (0)    & \makebox[0.05\textwidth][s]{28} (60.7)  \\
    Disruption     & \makebox[0.05\textwidth][s]{115} (93.9)  & \makebox[0.02\textwidth][s]{2} (100.0) & \makebox[0.05\textwidth][s]{6} (100.0)  & \makebox[0.05\textwidth][s]{2216} (\textbf{43.1})  \\
    Floor-grabber  & \makebox[0.05\textwidth][s]{257} (75.5)  & \makebox[0.02\textwidth][s]{0} (0)     & \makebox[0.05\textwidth][s]{0} (0)  & \makebox[0.05\textwidth][s]{1152} (\textbf{49.7})  \\
    Question       & \makebox[0.05\textwidth][s]{10} (20.0)   & \makebox[0.02\textwidth][s]{0} (0)     & \makebox[0.05\textwidth][s]{1231} (\textbf{8.9})        & \makebox[0.05\textwidth][s]{0} (0)  \\
    Statement      & \makebox[0.05\textwidth][s]{9445} (14.5) & \makebox[0.02\textwidth][s]{79} (\textbf{8.9})  & \makebox[0.05\textwidth][s]{51} (60.8)    & \makebox[0.05\textwidth][s]{2} (100.0)  \\
    \hline
    \end{tabular}
    %}
    \caption{Punctuation vs. dialog act counts for MRDA dataset. Percentage of errors for a given act and punctuation are shown in parenthesis (the lower, the better the recognition).}
    \label{tab:mrda_punc_vs_da}
\end{table*}

\subsection{The effect of punctuation -- SWDA}

\begin{table*}
    \centering
    \begin{tabular}{lrrrr}
    \hline
    \textbf{Mis-segmented dialog acts} &  \textbf{Count} &  \textbf{DSER (lc) [\%]} &  \textbf{DSER (nlc) [\%]} &  \textbf{Abs. gain [\%]} \\
    \hline
    Rhetorical-Questions         &     12 &              58.3 &                16.7 &          -41.7 \\
    Uninterpretable              &    366 &              42.9 &                 6.8 &          -36.1 \\
    Hedge                        &     23 &              39.1 &                 4.3 &          -34.8 \\
    Quotation                    &     18 &              66.7 &                44.4 &          -22.2 \\
    Other                        &     15 &              40.0 &                20.0 &          -20.0 \\
    Statement-non-opinion        &   1494 &              30.7 &                13.7 &          -17.0 \\
    Agree-Accept                 &    213 &              22.1 &                 7.0 &          -15.0 \\
    Statement-opinion            &    832 &              29.7 &                15.9 &          -13.8 \\
    Declarative-Yes-No-Question  &     38 &              18.4 &                 5.3 &          -13.2 \\
    Open-Question                &     16 &              12.5 &                 0.0 &          -12.5 \\
    \hline
    \end{tabular}

    \caption{Top 10 SWDA dialog acts that benefit from punctuation and truecasing availability in pure dialog act segmentation. The columns denoted by (l) and (nl) represent numbers for dialog-level context XLNet lower and nolower models, respectively.
    }
    \label{tab:punctuation_diff}
\end{table*}

\begin{table}
    \centering
    \resizebox{\columnwidth}{!}{
    \begin{tabular}{lrrrr}
        \hline
        \textbf{Segmentation} &    \textbf{Full stop} & \textbf{Comma} & \textbf{Q. mark} & \textbf{Segments} \\
        \hline
        ground truth    &   71 &  3637 &  2 &      4500 \\
        nolower         &   77 &  3679 &  2 &      4433 \\
        lower           &  155 &  3737 &  7 &      4323 \\
        \hline
    \end{tabular}
    }
    \caption{The number of punctuation symbols found in the middle of dialog acts, depending on the applied segmentation. \textit{nolower} and \textit{lower} are predicted using XLNet with dialog-level context. The presence of punctuation in \textit{nolower} variant provides the model with the necessary cues to preserve a similar distribution to the ground truth.}
    \label{tab:model_type_vs_punctuation_in_segments}
\end{table}

\begin{figure}[h!]
    \centering
    \resizebox{\columnwidth}{!}{
    \includegraphics{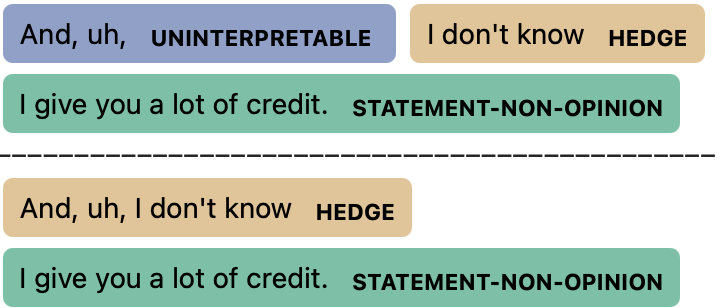}
    }
    \caption{Top: ground truth segmentation. Bottom: segmentation predicted with \emph{lower} transcripts.}
    \label{fig:uninterpretable_example}
\end{figure}

Given the large label set size of SWDA, we have no straightforward means of visualizing the correlation of punctuation and dialog acts. In order to understand the relationship between punctuation and dialog acts in SWDA, we show the top 10 most affected dialog acts in segmentation and recognition in Table~\ref{tab:punctuation_diff}. We observe that punctuation is key in recognizing discourse markers such as incomplete utterances, restarts, or repairs that are often labeled as \emph{Uninterpretable}. Without punctuation, these discourse markers are frequently merged into a neighboring dialog act by the model. It also partially explains the improvements in segmentation of \emph{Statements} and some less frequent acts such as \emph{Hedge}, since they are often found next to \emph{Uninterpretable} (see Figure~\ref{fig:uninterpretable_example}).

\begin{figure}[h!]
    \centering
    \resizebox{\columnwidth}{!}{\includegraphics{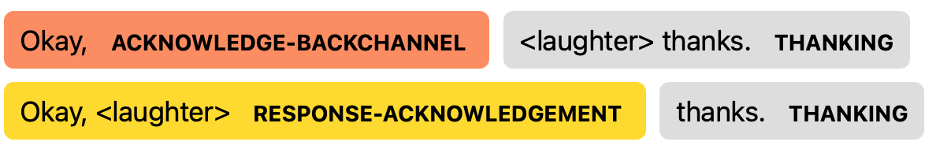}}
    \resizebox{\columnwidth}{!}{\includegraphics{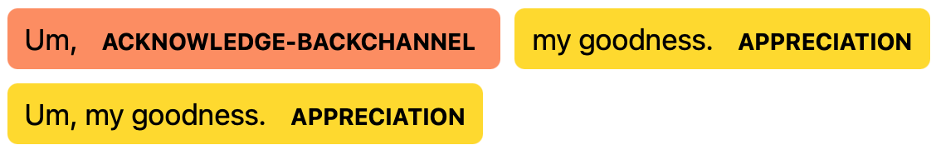}}
    % \resizebox{\columnwidth}{!}{\includegraphics{figures/no_kidding_really.png}}
    \caption{Top: ground truth segmentation. Bottom: segmentation predicted with \emph{lower} transcripts.}
    \label{fig:okay_thanks_example}
\end{figure}

In many cases, the lack of commas takes away a cue to insert a dialog act boundary from the model. Examples are shown in Figure~\ref{fig:okay_thanks_example}. We hypothesize that prosody or other cues found in the acoustic signal could mitigate that effect, given the usefulness of such features in dialog act classification works~\cite{ortega2018lexico,si2020hierarchical}.

Another way to look at the differences in the segmentation structure is to compare the distributions of punctuation symbols found in the middle of the segments (i.e., the punctuation symbols other than the ones ending the previous and the current dialog act). We present them in Table~\ref{tab:model_type_vs_punctuation_in_segments}. We see that the \emph{nolower} model uses the punctuation as cues for determining segment boundary and retains a very similar distribution to the ground truth segmentation. On the other hand, the \emph{lower} model, which cannot see the punctuation, tends to under-segment the transcripts. This is consistent with our previous analyses.

\section{Conclusions}
\label{sec:conclusions}

We investigated how two transformer models capable of dealing with long sequences, XLNet and Longformer, can be applied to dialog act recognition. We used the well-studied SWDA and MRDA corpora and compared the performance with an established BiGRU baseline. First, we showed that the pre-trained transformers offer a substantial improvement w.r.t. to BiGRU when processing individual speaker turns, without any additional context. Then, we proposed adapting the transformers to consider a broader dialog context through turn concatenation with the \textit{TURN} token, the use of joint coding, and local attention patterns or windowed processing. With this improvement, we achieved strong segmentation results on SWDA and MRDA dialog act recognition with DSER of 8.4\% and 14.2\% on the original transcripts and competitive results on lowercase transcripts with no punctuation (17.5\% and 32.9\%). 

We found that XLNet was able to get the most out of the additional dialog context. We observed that the additional context is the most beneficial for segmentation while also improving the classification performance. Of a practical note, Longformer allowed for approximately five times quicker inference on a modern CPU. 

Across all of our experiments, it was evident that punctuation and original character cases were crucial for both segmentation and classification performance. No other factor influences the results as much - the best lowercase-transcript model (broad context XLNet\textit{+prop}) still lags behind the simplest unmodified-transcript model (turn-context BiGRU). 
We analyzed the effect of punctuation and found that it is often correlated with some dialog act classes. 
The model leverages punctuation as a cue, especially to insert segment boundaries, but to a lesser extent also to classify dialog acts (e.g., question marks in questions). 

By considering different dialog act label sets available in MRDA and a pure segmentation task, we found that XLNet's segmentation performance does not depend on the dialog act labels, further with segmentation experiments on SWDA. Regardless of the label set size (or whether the task is pure segmentation), the model performs just as well.  

Finally, we found that the addition of broader context is beneficial for the model to learn rare dialog act classes -- without it, more than 50\% of dialog act classes were never correctly recognized even once in SWDA. With the inclusion of context, that number decreased to less than 10\%.

Our findings have significant practical implications for applications that depend on text segmentation, such as the automatic discovery of intents and processes in a given domain or building graphs describing conversational flow from unstructured transcripts. We have shown that the dialog act labels do not have to be specific in order to be able to retrieve good segmentation automatically. This can significantly ease the annotation efforts, removing the need to memorize large label sets for the annotators. Furthermore, we show that the current pre-trained transformer models suffer from limitations when punctuation is not available. They tend to under-segment the text, often merging disfluencies with neighboring dialog acts. While these phenomena would likely affect, e.g., systems trying to measure the semantic similarity of two segments, we expect that even the segmentation predicted on lower-case text would be useful in practical applications. It is interesting to see whether automatically retrieved punctuation can mitigate the gap between manual annotation and no punctuation; we consider this a promising future work candidate.

To foster further research in this direction, we make our code available under the Apache 2.0 license\footnote{\url{https://github.com/pzelasko/daseg/tree/version/tacl2021}}.

\bibliography{tacl2018}
\bibliographystyle{acl_natbib}

\end{document}